\documentclass{bmvc2k}

\usepackage{amsmath,amssymb,amsfonts}
\usepackage{algorithmic}
\usepackage{graphicx}

\usepackage{amsmath}

\DeclareMathOperator*{\argmin}{arg\,min}

\title{Curriculum based Dropout Discriminator for Domain Adaptation}

\addauthor{Vinod Kumar Kurmi}{vinodkk@iitk.ac.in}{1}
\addauthor{Vipul Bajaj}{vipulbjj@iitk.ac.in}{1}
\addauthor{Venkatesh K Subramanian}{venkats@iitk.ac.in}{1}
\addauthor{Vinay P Namboodiri}{vinaypn@iitk.ac.in}{1}

\addinstitution{
 Indian Institute of Technology Kanpur\\
 Kanpur, India
}
\runninghead{V.K. Kurmi, V. Bajaj, K.S. Venkatesh, V.P. Namboodiri}{C$\text{D}^{3}$A}

\begin{document}

\maketitle

\begin{abstract}
Domain adaptation is essential to enable wide usage of deep learning based networks trained using large labeled datasets. Adversarial learning based techniques have shown their utility towards solving this problem using a discriminator that ensures source and target distributions are close. However, here we suggest that rather than using a point estimate, it would be useful if a distribution based discriminator could be used to bridge this gap. This could be achieved using multiple classifiers or using  traditional ensemble methods. In contrast, we suggest that a Monte Carlo dropout based ensemble discriminator could suffice to obtain the distribution based discriminator. Specifically, we propose a curriculum based dropout discriminator that gradually increases the variance of the sample based distribution and the corresponding reverse gradients are used to align the source and target feature representations. The detailed results and thorough ablation analysis show that our model outperforms state-of-the-art results.
\end{abstract}

\section{Introduction}
Visual recognition has seen vast improvements based mainly on the success of deep learning based models ~\cite{krizhevsky_NIPS2012}. These models are trained on very large annotated datasets such as Imagenet~\cite{ILSVRC15}. The deployment of these generically trained models require them to adapt to work in specific settings (for instance with catalog images in E-commerce websites). This problem is recognized as one of dataset bias and was demonstrated through the work of~\cite{torralba_CVPR2011}. However, the requirement of a large annotated dataset becomes a bottleneck for training networks in deep learning frameworks. 
In this paper, we tackle the problem of adapting classifiers to work on datasets that \emph{do not} have any labeled information. This problem is one of unsupervised domain adaptation in a more general setting.


Ganin and Lempitsky~\cite{ganin_ICML2015} proposed a method to solve unsupervised domain adaptation through back-propagation. In this method, the domain adaptation problem is solved by using a discriminator that ensures domain invariance of learned representations used for classification. There have been several methods \cite{tzeng_CVPR2017, hoffman_arxiv2017,shen_arxiv2017} proposed for improving the discriminator. However, most of these involve an increase in the number of parameters. For instance a recent work by Pei \textit{et al} (MADA)~\cite{pei_arxiv2018} addresses this issue through class-specific discriminators. This leads to a linear increase in the number of parameters with the number of classes in  dataset. In contrast, we propose the use of curriculum-based dropout discriminator to obtain improved performance of the domain adaptation task without increasing the number of parameters. It makes our model's applicability comprehensive as it can also adapt to datasets with a large number of classes.
\begin{figure}
 \centering
    \includegraphics[height=6cm,width=12.5cm]{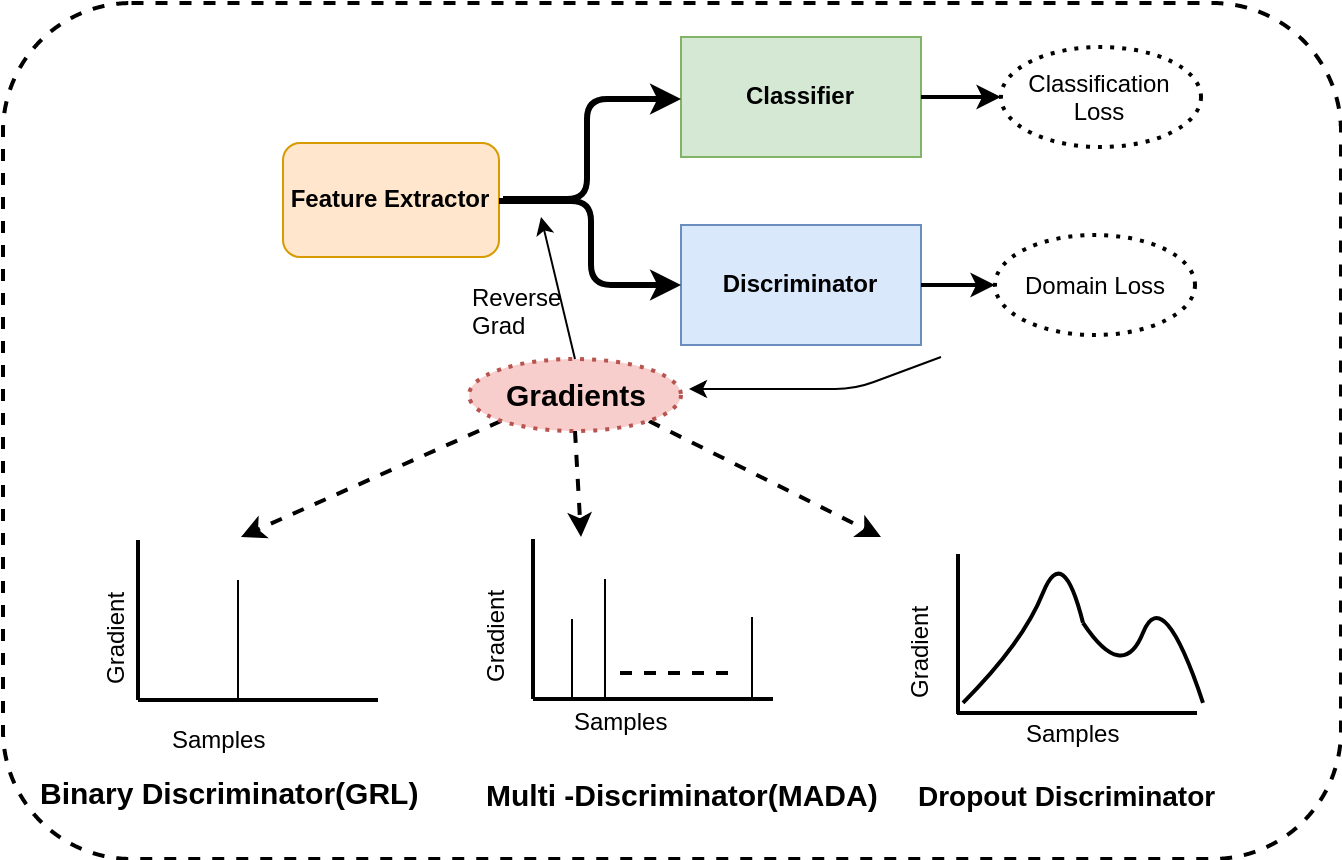}
      \caption{The difference in adversarial learning framework for domain adaption using binary discriminator, multi-discriminator and dropout discriminator. In binary discriminator~\cite{ganin_ICML2015}, the feature extractor is trained with one gradient value of the discriminator. In the case of multi-discriminator~\cite{pei_arxiv2018}, learning occurs with a fixed number of gradient values, whereas in dropout based discriminator, feature extractor learns a distribution rather than a single value. }
      \label{fig:intro}
      \vspace{-1.2em}
 \end{figure}
Specifically, in this paper, we propose Curriculum based Dropout Discriminator for Domain Adaptation (\textbf{C$\text{D}^{3}$A}) and compare it with a variant, Dropout Discriminator for Domain Adaptation (\textbf{$\text{D}^{3}$A}).  It's a novel approach that solves the above problem through an adversarial dynamic dropout based ensemble of discriminators,  where we consider dropout as being a source of an ensemble of domain classifiers \cite{hara2016analysis}. The proposed model also enables the discriminators reduce the prediction variance, remove overfitting, and average out the bias. The idea for this discriminator is illustrated in Figure~\ref{fig:intro}. The initial discriminator by Ganin and Lempitsky \cite{ganin_ICML2015} suggests the use of a single binary discriminator and MADA~\cite{pei_arxiv2018} extends it to class-specific cues. 
In contrast, C$\text{D}^{3}$A obtains a discriminator distribution that provides a much-improved feedback for improving the feature extractor. The performance of any adversarial learning method largely depends upon the capability of the discriminator network. The ensemble method~\cite{hara2016analysis} improves the discriminator's performance and makes it robust. We show that this indeed helps in an improved domain adaptation (around 5.3\% improvement in Amazon-DSLR adaptation) with much fewer parameters($\sim
$59M) than MADA($\sim$98M). More importantly, our method does not increase the number of parameters as the number of classes increase, making it scalable to datasets with a large number of classes. Through this paper we make the following main contributions:


\begin{itemize}
\item We propose a method to obtain a dropout based discriminator that provides a distribution based discrimination for every sample ensuring a more robust feature adaptation
\item We adopt a curriculum based dropout model, C$\text{D}^{3}$A, that ensures gradual increase in the number of samples as the adaptation progresses to ensure better adaptation in contrast to a fixed number of samples based dropout distribution (D$^{3}$A).
\item We provide a thorough empirical analysis of the method (including statistical significance, discrepancy distance) and evaluate our approach against the state-of-the-art approaches.
\end{itemize}

\section{Related Work}


\qquad \textbf{{Domain Adaptation:}}
 A large number of methods have been proposed to tackle the domain adaptation problem. The basic common structure that has been followed is the Siamese architecture~\cite{bromley_NIPS1994} with two streams, representing the source and target models. It is trained with a combination of a classification loss and the other being one of discrepancy loss or an adversarial loss. The classification loss depends on the source data label,  while the discrepancy loss reduces the shift between the two domains. A discrepancy based deep learning method is that of deep domain confusion (DDC)~\cite{tzeng_arxiv2014}. The loss between a single FC (fully connected) layer of source and target feature extractor network is used to minimize the maximum mean discrepancy (MMD) between the source and the target. This approach is further extended by deep adaptation network (DAN)~\cite{long_ICML2015}. 
 Recently, a number of other methods have been proposed which use discrepancy of domain~\cite{saito_cvpr2017maximum,zhang_cvpr2018aligning,sun_ECCV2016,sun_DACVA2017,sun_2016AAAI,shen_AAAI2018wasserstein,long_ICML2017,rozantsev_PAMI2018}.

\textbf{{Adversarial Learning:}}
In the domain adaptation setting, an adversarial network provides domain invariant representations by making the source and target domain indistinguishable by the discriminator. Adversarial Discriminative Domain Adaptation~\cite{tzeng_CVPR2017} uses an inverted label GAN loss to split the optimization into two independent objectives. One such method is the domain confusion based model proposed in~\cite{tzeng_ICCV2015} that considers a domain confusion objective. Domain-Adversarial Neural Networks (DANN)~\cite{ganin_ICML2015} integrates a gradient reversal layer into the standard architecture to promote the emergence of the learned representations that are discriminative for the main learning task on the source domain and non-discriminative concerning the shift between the domains. Recently, some works have been proposed which use an adversarial discriminative approach in solving the domain adaptation problem~\cite{saito2018adversarial,hoffman_2018cycada,bousmalis_2017CVPR,zhang_cvpr2018importance,chen_cvpr2018re,li_cvpr2018domain,kurmi2019attending}. 
Similarly, the model proposed in~\cite{bousmalis_CVPR2017,choi2017_cvprstargan} exploits GANs with the aim to generate source-domain images such that they appear as if they were drawn from the target domain distribution. The closest related work to our approach is the work by~\cite{pei_arxiv2018} that extends the gradient reversal method  by a class-specific discriminator.

\textbf{{Ensemble and Curriculum learning:}}
Ensemble methods~\cite{lakshminarayanan_nips2017} can capture the uncertainty of the neural network (NN). Gal \textit{et.al}.~\cite{gal_icml2016dropout} use dropout to obtain the predictive uncertainty and apply Markov chain Monte Carlo~\cite{neal_2012bayesian} also known as MCMC at the test time to deal with intractable posterior. 
In discriminator based approaches, ensembles can be considered as multi-discriminator or multi-generator architecture. Multi discriminator approach has also been proposed by~\cite{nguyen_nips2017dual,ghosh_corr2017multi,durugkar_2016generative} to learn the data distribution more effectively. In Bayesian GAN~\cite{saatci_nips2017bayesian}, dropout in the discriminator is used which can be interpreted as an ensemble model~\cite{gal_icml2016dropout}. The curriculum learning~\cite{bengio2009curriculum} enhances model's performance and its generalization capability. The performance of the GAN is also improved through the curriculum learning of the discriminator~\cite{sharma2018improved}. It has been shown that dropout can also work with curriculum learning~\cite{morerio2017curriculum}. In domain adaptation, a curriculum style learning approach has been applied in ~\cite{zhang2017curriculum} to minimize the domain gap in semantic segmentation. The curriculum domain adaptation first solves easy tasks such as estimating label distributions, then infers the necessary properties about the target domain. The theoretical framework for curriculum learning in transfer learning is proposed in~\cite{weinshall2018curriculum}. Recently other curriculum learning based domain adaptation methods have been proposed in Transferable Curriculum Learning~\cite{shu2019transferable}.  

In contrast to the previous works, the main contribution of the present work is to propose a curriculum based dropout discriminator. We show that through the proposed method, we are able to outperform state of the art domain adaptation techniques in a scalable way by using fewer number of parameters as compared to techniques such as MADA\cite{pei_arxiv2018} and similar number of parameters as GRL \cite{ganin_ICML2015}.

\section{Motivation}
In the adversarial domain adaptation problem, the previous methods have used classical statistical inference in the discriminator. A single discriminator learns the source and target domain classification. Our hypothesis is that it may lead to overconfident inference and decisions which in turn may lead to challenges in learning invariant features. In the domain adaptation problem, data is generally structured in a multimodal distribution. Thus, a multiple discriminator approach is compelling~\cite{pei_arxiv2018}, due to its capacity to capture multiple modes of the dataset. It also leads to solving the perennial problem of mode collapse (which GANs are infamous for) as multiple discriminators now learn to distinguish classes with different modes. The diversity of an ensemble of such discriminators reduces the random errors in prediction. The performance of an ensemble model rests on the number of entities in the ensemble. However, as the number of entities increase, the model parameters and complexity will increase. This is one of the primary bottlenecks of the ensemble based methods. The number of parameters in an algorithm is a significant factor in determining model efficiency.

To tackle the above problems, we propose a novel and efficient discriminator architecture by using Monte Carlo (MC) sampling~\cite{srivastava_jmlr2014dropout}. We incorporate Bernoulli dropout in a multi-adversarial network, by dropping out a certain number of neurons from our discriminator with some probability \textit{d}. This gives rise to a set of dynamic discriminators for every data sample. The main idea behind our method is to construct a training regime for the feature extractor in domain adaptation that consists of increasingly challenging tasks to generate domain invariant features. This allows the sophistication of the feature extractor to gradually increase throughout training, rather than aiming for full sophistication at the outset. This method is similar to that of curriculum in supervised learning, where one orders the training examples to be presented to a learning
algorithm according to some measure of difficulty~\cite{bengio2009curriculum}. Despite the conceptual similarity, the methods are quite different. Under our approach, it is not the difficulty of the training examples presented to either network, but rather the \textit{capacity}, and hence strength, of the discriminator network that is increased as the training progresses. The idea behind the use of a curriculum based dropout discriminator is to exploit the characteristics of several independent discriminators by consolidating them in order to achieve higher performance.

We do a curriculum based learning on these dropout discriminators. As the training proceeds, the number of discriminators sampled, increase, thereby boosting the variance of our model's prediction. The proposed approach enforces the feature extractor network not to constrain the learned representations to satisfy a single discriminator, but, instead, to satisfy an ensemble of dynamic discriminators (composition is different across different discriminators). Instead of learning a point estimate (in case of MADA~\cite{pei_arxiv2018}), the feature extractor network of our proposed model learns a distribution, due to the ensemble effect of feedback from a set of dynamic discriminators. This approach leads to a more generalized feature extractor, promoting resemblance in learned representations of a class from different domains.  The instinct behind incorporating dropout in our model is to warrant that neurons are not exclusively reliant on a precise set of other neurons to determine their outputs. Instead, each neuron relies on the agglomerate behavior of several other neurons, promoting generalization. By applying dropout on the discriminator, we obtain a set of entirely dynamic discriminators and hence the feature extractor cannot use the trick of relying on a specific type of discriminator or ensemble of discriminators to learn to generate representations to deceive the discriminator. Instead, it will now have to genuinely learn domain invariant representations. Thus, the feature extractor network is now guided by diverse feedback given to it by an ensemble of dynamic discriminators.
All this increase in performance is obtained without compromising on the scalability and complexity front through our proposed model.

 

\begin{figure}
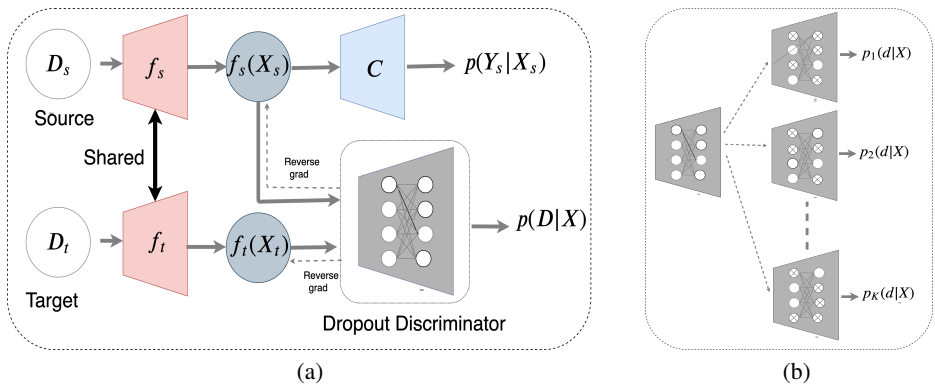

     \small
     \begin{tabular}[b]{ c c }
       \includegraphics[height=4.5cm,width=8cm]{main_fig_1.png}
     & \includegraphics[height=4.5cm,width=4cm]{drop_fig.png}\\
     (a) & (b)
     
       \end{tabular}
      \caption{(a) Proposed model includes the source and target feature extractor(shared), classifier network using the fully connected layers and the dynamic ensemble of discriminators using the Bernoulli dropout network (b) Dropout based discriminator architecture }
      \label{fig:main}
 \end{figure}
\section{Proposed Adaptation Model}
In the unsupervised domain adaptation problem, we consider that the \textit{source dataset} $\mathcal{D}_s$ has access to all its labels while there are no labels for the \textit{target dataset} $\mathcal{D}_t$ at the training time. We assume that $\mathcal{D}_s$ comes from a source distribution $ \mathcal{S}$ and $\mathcal{D}_t$ comes from a target distribution $ \mathcal{T}$. We assume that there are $N_s$ source data points and $N_t$ unlabeled target data points.
So $ \mathcal{D}_s = {(x_i^s,y_i^s)}_{i=1}^{N_s} \in   \mathcal{S}$ has $N_s$ labeled examples and the target
domain $\mathcal{D}_t = {(x_i^t)}_{i=1}^{N_t} \in  \mathcal{T} $  has $N_t$ unlabeled examples.
Our underlying assumption is that both distributions are complex and unknown.
Our model provides a deep neural network that enables learning of transferable feature representations $f(x)$ and an adaptive  classifier $y = C(f(x))$ to reduce the
shift in the joint distributions across domains, such that the
target risk $Pr_{(x,y)\sim q }[C(f(x))\neq y]$ is minimized by jointly
minimizing source risk and distribution discrepancy by adversarial 
domain adaptation where $q$  is assumed to be the joint distribution of target samples.

In this work, we employ a variant of GRL\cite{ganin_ICML2015}, where discriminator is modeled as an MC-dropout based ensemble. The feature extractor network consists of convolution layers to produce image embeddings. Both source and target feature extractors share the same parameters. The classifier network consists of fully connected layers. Only source embeddings are forwarded to the classifier network to predict the class label. The classifier network parameters ($\theta_c$) are updated only by the loss from source data samples. The discriminator receives both source and target embeddings. The parameters of the MC-dropout discriminator ($\theta_d$) are updated with domain classification loss. The feature extractor parameters($\theta_f$) are updated by the gradients from the classifier network as well as by the reverse gradient of both source and target data samples from the dynamic set of the ensemble of discriminators. Detailed architecture is presented in Figure~\ref{fig:main}.

For the adaptation task, the feature extractor learns domain-invariant features with the help of MC-Dropout based discriminator. For each data sample that goes to the discriminator, we obtain the domain classification loss. These losses are backpropagated through respective Monte Carlo sampled dropout discriminators followed by gradient reversal layer. Hence, for every input, we obtain a distribution of gradients. The feature extractor is updated by a gradient from this distribution to generate domain invariant features. In a binary discriminator~\cite{ganin_ICML2015}, we obtain a point estimate of the gradient for specific input. In the case of multi discriminator~\cite{pei_arxiv2018}, we obtain an ensemble of the point estimates of gradients. The advantage of obtaining a distribution of gradients is that we get generalized learned representations robustly leading to domain invariant features. We propose Curriculum based Dropout Discriminator (C$\text{D}^{3}$A), where we increase the number of MC samples as training proceeds in a paradigm similar to curriculum learning. However, in the other variant($\text{D}^{3}$A), we maintain a fixed number of MC sampled discriminators throughout the training.
\vspace{-1.45em}
\subsection{Curriculum based Dropout Discriminator for Domain Adaptation (C$\text{D}^{3}$A)}
 In C$\text{D}^{3}$A, the distribution of gradients is obtained through a curriculum fashioned training, i.e., we increase the number of MC samples as training proceeds. The motivation behind increasing the number of MC samples is that, in the initial phase of the adaptation, the feature extractor learns the domain invariant features without considering the multi-mode structure of data. For this purpose, only a small number of discriminators is required. As the training advances, we expect the network to learn the domain invariant features along with its multi-modal structure. Thus, in the proposed model, we increase the MC samples of discriminator as training progress to obtain the domain invariant feature without losing its multi-mode structure. 
Given an input sample $x_i$, we obtain feature embedding $f(x_i)$, by passing it through a feature extractor $f$. These embeddings are further used to obtain the classification score $C(f(x_i))$ and the domain classification score for $j^{th}$ samples of discriminator $D_j(f(x_i))$, where $j=1,..., K$.
The curriculum learning of the discriminator does not rely on the difficulty of the training examples presented to either network, but rather the capacity, and hence strength, of the discriminator that is increased throughout the training. We construct an ordered set of sets of samples of discriminator increasing in numbers. More formally the set of discriminators is
$\mathcal{D}=\{ \{ D_1\},\{ D_1,D_2\}, ...,\{ D_1,D_2, .. D_K\} \} $, where $D_j$ is a MC sampled discriminator. We can clearly see that the $\mathcal{D}$ is a ordered set in terms of the capacity, where capacity of $\{ D_1\} \subseteq \{ D_1,D_2\}$.
\vspace{-1em}
\subsection{Fixed sampling based Dropout Discriminator for Domain Adaptation($\text{D}^{3}$A)}
In this variant, we fix the number of MC sampled discriminators during the training. In this scenario, we obtain an ensemble of discriminators. We call this variant as a Dropout Discriminator for Domain Adaptation( $\text{D}^{3}$A). This modification can be considered a more efficient version of the multi discriminator model. We experimented with different sampling values (details are reported in supplementary) and obtained the best results when the number of samples is chosen close to the number of classes in the target dataset. 



\subsection{Loss Function}
Our loss function is composed of classification loss and domain classification loss. Our classifier takes learned representations as input and predicts its label.  Classification loss function $\mathcal{L}_c$ is a cross-entropy loss. Dropout discriminator is expected to label (output) the source domain images as 0 and target domain images as 1. Domain classification loss $\mathcal{L}_d$ is a binary cross entropy loss between the output of discriminator and the expected output. It is summed over the number of MC-sampled discriminators~$K$.  $K$ is increased as the training proceeds in the case of CD$^3$A model, whereas it is fixed for D$^3$A model. 
\begin{equation}
\begin{split}
\mathcal{L}(\theta_f,\theta_c, \theta_d) = & \frac{1}{N_s}\sum_{x_i \in \mathcal{D}_s} \mathcal{L}_c(C(f(x_i)),y_i)   - \frac{\lambda}{N}\sum_{j=0}^{K} \sum_{x_i \in \mathcal{D}_s \cup \mathcal{D}_t} \mathcal{L}_d(D_j(f(x_i)),d_i)
\end{split}
\end{equation} 

\begin{equation}
    (\hat{\theta}_f,\hat{\theta}_c)= \argmin_{\theta_f,\theta_c}  \mathcal{L}(\theta_f,\theta_c,\hat{\theta_d})  \hspace{2em}  \hat{\theta_d}= \argmin_{\theta_d} \mathcal{L}(\hat{\theta_f},\hat{\theta_c},\theta_d)
\end{equation}


where $ d_i=0$ $\text{if $x_i \in \mathcal{D}_s $}$ and $ d_i=1$ $\text{if $x_i \in \mathcal{D}_t $}$ . The function $f$ is the feature extractor network with shared weights for source and target data ($f_s$ and $f_t$ are denoted by common shared network $f$). $\lambda$ is the trade-off parameter between the two objectives. $C$ is the classifier network and $D_j$ is the $j^{th}$ MC-sampled dropout discriminator. $\mathcal{D}_s$ and $\mathcal{D}_t$ represent source and target domains respectively.

We generate the entities of ensemble via dropout. In contrast, previous works~\cite{pei_arxiv2018} use multiple discriminators; their number being equal to the number of classes in the dataset. It leads to an increase in the number of parameters employed in the discriminator which makes it unsuitable for datasets with a large number of classes. Also, due to our model's parameters being significantly less, the data requirements are also quite low. This has been shown in supplementary material, where we remove half of the source data and still obtain good accuracy. Also, MADA uses the predicted label probabilities to weigh the discriminator's response. This is a drawback as it can lead to misleading corrections of the feature extractor network in case of wrong predictions by the label predictor (classifier). Our model doesn't have such constraints making our discriminator even more powerful leading to better learning of domain invariant features by the feature extractor network. 
The implementation details are provided in the supplementary material, and other details are provided on the project page~\footnote{https://delta-lab-iitk.github.io/CD3A/}.

\section{ Results and Analysis}
\subsection{Datasets}
 \textbf{{Office-31 Dataset:}}
Office-31~\cite{saenko_ECCV2010} is a benchmark dataset for domain adaptation, comprising 4,110 images in 31 classes collected from three distinct domains: Amazon (A), Webcam (W) and DSLR (D). To enable unbiased evaluation, we evaluate all the methods on all 6 transfer tasks A $\rightarrow$ W, D $\rightarrow$ A, W $\rightarrow$ A, A $\rightarrow$ D , D $\rightarrow$ W and W $\rightarrow$ D.
\newline
\textbf{{ImageCLEF Dataset:}}
ImageCLEF-2014 dataset consists of 3 domains: Caltech-256 (C), ILSVRC 2012 (I), and Pascal-VOC 2012 (P). There are 12 common classes, and each class has 50 samples.There is a total of 600 images in each domain. We evaluate models on all 6 transfer tasks: I$\rightarrow$P, P$\rightarrow$I, I$\rightarrow$C, C$\rightarrow$I, C$\rightarrow$P, and P$\rightarrow$C. 

 \begin{table*}[!h]
 \begin{center}
\begin{tabular}{ |c|c|c|c|c|c|c|c| }
 \hline
  \textbf{Method }& A $\rightarrow$ W & D$\rightarrow$ W &  W $\rightarrow$ D &A $\rightarrow$ D & D $\rightarrow$ A & W $\rightarrow$ A & Average \\ 
  \hline
 Alexnet~\cite{krizhevsky_NIPS2012}  & 60.6 & 95.0   & 99.5 &64.2 & 45.5  & 48.3 & 68.8\\
  MMD\cite{tzeng_arxiv2014} & 61.0  & 95.0  & 98.5 &64.9 & 47.2 & 49.4&69.3 \\ 
  RTN\cite{long_NIPS2016} & 73.3  & 96.8  & 99.6& 71.0& 50.5 & 51.0 & 74.1\\ 
  DAN\cite{long_ICML2015} & 68.5 & 96.0  & 99.0  & 66.8 & 50.0 & 49.8 & 71.7 \\ 
  GRL \cite{ganin_ICML2015} & 73.0 & 96.4  & 99.2  & 72.3  & 52.4  & 50.4 & 74.1\\
  
  JAN \cite{long_icml2017deep} & 75.2  & 96.6  & 99.6  & 72.8  & 57.5  & 56.3 & 76.3\\
  CDAN\cite{long_arxive2017conditional} & 77.9  & 96.9  &  100.0  & 74.6  & 55.1  &  \textbf{57.5} & 77.0\\ 
 MADA\cite{pei_arxiv2018} & 78.5  & {99.8 } &  100.0  & 74.1  & 56.0 & 54.5 & 77.1\\ 
 IDDA\cite{kurmi2019looking} & 82.2 & 99.8 & 100.0 & \textbf{82.4} & 54.1 & 52.5& 78.5 \\ 
 \hline
   \textbf{$D^{3}$A}(31)& 79.0 & 97.7 & {100.0 } & {79.4 } & {58.2}  & {55.3} & { 78.3} \\
   \hline
 \textbf{CD$^{3}$A}& \textbf{{82.3}} &\textbf{99.8} &\textbf{100.0} & { 81.1}  &\textbf{ 58.2} & 55.6 & \textbf{79.5} \\ 
 \hline
 \end{tabular}
\end{center}
\caption {Classification accuracy (\%) on Office-31 dataset for unsupervised domain adaptation on AlexNet\cite{krizhevsky_NIPS2012} pretrained network. Our model is C$D^{3}$A  and $D^{3}$A with the number in bracket indicating the number of Monte Carlo samples \label{table:office_table}} 
 \end{table*}
\begin{table*}[!h]
\centering
\begin{tabular}{ |c|c|c|c|c|c|c|c| }
 \hline
  \textbf{Method }& I$\rightarrow$P & P$\rightarrow$I &  I$\rightarrow$C &C$\rightarrow$I & C$\rightarrow$P & P$\rightarrow$C & Avg \\ 
 \hline
AlexNet~\cite{krizhevsky_NIPS2012} & 66.2 & 70.0 & 84.3 & 71.3 & 59.3 &84.5 & 73.9  \\
DAN\cite{long_ICML2015} & 67.3 & 80.5 & 87.7 & 76.0 & 61.6 & 88.4 & 76.9  \\
GRL~\cite{ganin_ICML2015} & 66.5 & 81.8 & 89.0 & 79.8 &63.5 &88.7 &78.2  \\
RTN\cite{long_NIPS2016} & 67.4 & 82.3 &89.5 & 78.0 &63.0 &90.1 &78.4  \\
MADA~\cite{pei_arxiv2018} & 68.3 &\textbf{83.0} &91.0 &80.7 &63.8 & \textbf{92.2} &79.8  \\
 \hline
\textbf{$D^{3}$A(12)} & {69.1} & {80.9} &  {91.0} & {81.5} &  \textbf{66.2}  &  {90.0} & {79.8} \\
 \hline
 \textbf{CD$^{3}$A}&\textbf{69.3}& 81.5 & \textbf{91.3} &\textbf{ 81.6} &{ 65.9} & {{90.2}} & \textbf{80.0}\\ 
 \hline
\end{tabular}
\vspace{1em}
\caption {Classification accuracy (\%) on \textit{ImageCLEF} dataset for unsupervised domain adaptation (AlexNet~\cite{krizhevsky_NIPS2012})\label{tbl:imageclef}} 
 \end{table*}

\begin{table*}[!]
\centering
\begin{tabular}{|c|c|c|c|c|c|c|c|}
 \hline
  \textbf{Method }& I $\rightarrow$ P & P$\rightarrow$ I &  I $\rightarrow$ C &C $\rightarrow$ I & C $\rightarrow$ P & P $\rightarrow$ C & Average \\ 
  \hline
ResNet~\cite{he2016deep} & 74.8 & 83.9 & 91.5 & 78.0 & 65.5 & 91.2 & 80.7 \\
DAN~\cite{long_ICML2015} & 75.0 & 86.2 & 93.3 & 84.1 & 69.8 & 91.3 & 83.3 \\
RTN~\cite{long_NIPS2016} & 75.6 & 86.8 & 95.3 & 86.9 & 72.7 & 92.2 & 84.9 \\
GRL~\cite{ganin_ICML2015} & 75.0 & 86.0 & 96.2 & 87.0 & 74.3 & 91.5 & 85.0 \\
JAN~\cite{long_ICML2017} & 76.8 & 88.0 &94.7 & 89.5 & 74.2 & 91.7 & 85.8 \\ 
MADA~\cite{pei_arxiv2018} & 75.0 & 87.9 & {96.0} & 88.8 & 75.2 & 92.2 & 85.8 \\
CDAN~\cite{long_arxive2017conditional} & 77.2 & 88.3 & 98.3 & 90.7 & 76.7 & 94.0 & 87.5  \\
 \hline
\textbf{CD$^{3}$A} & \textbf{77.5} & \textbf{88.7} & \textbf{96.8} & \textbf{93.2} &  \textbf{78.3}  &  \textbf{94.7} & \textbf{88.2} \\
 \hline

\end{tabular}
\vspace{1em}
\caption {Classification accuracy (\%) on \textit{ImageCLEF} dataset for unsupervised domain adaptation (ResNet-50~\cite{he2016deep})\label{tbl:imageclef_res}} 

\end{table*}

\subsection{Results}

We use pre-trained Alexnet~\cite{krizhevsky_NIPS2012} architecture following the typical setting in unsupervised domain adaption for our base model. Table~\ref{table:office_table} summarizes results on Office31 dataset,
and Table~\ref{tbl:imageclef} and Table~\ref{tbl:imageclef_res} have results for the ImageClef dataset for AlexNet and ResNet networks respectively. The results on Office-Home~\cite{venkateswara_cvpr2017deep} along with the implementation details are provided in the supplementary material. We obtained state-of-the-art results on all the datasets. It is noteworthy that the proposed model boosts the classification accuracies substantially on hard transfer tasks, e.g., A $\rightarrow$D, A $\rightarrow$W, etc. where the source and target domains are substantially different. On average, we obtain considerably improved accuracies and statistically significant results as shown further.

\begin{figure*}[!]
     \small
     \centering
     \begin{tabular}[b]{ c  c c c  }
     (a) t-SNE plot of &  (b) t-SNE plot of & (c)Proxy-Distance &  (d)Proxy-Distance    \\ RevGrad &  C$D^{3}$A &   A$\rightarrow$D &  A$\rightarrow$W  \\ 
     \includegraphics[scale=0.30, width=0.20 \linewidth, height = 0.1 \textheight ]{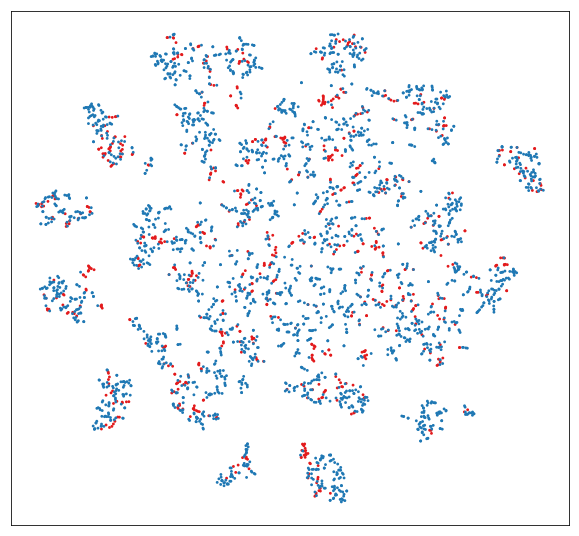}
     & \includegraphics[scale=0.30, width=0.20 \linewidth ,height=0.10 \textheight]{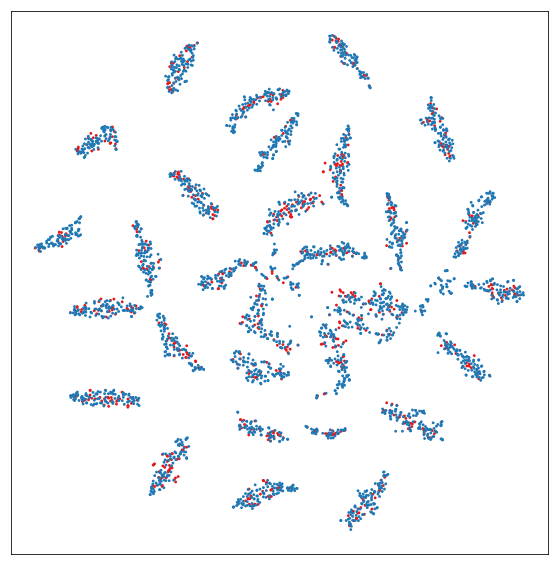}
     & \includegraphics[scale=0.30, width=0.20 \linewidth ,height=0.10 \textheight]{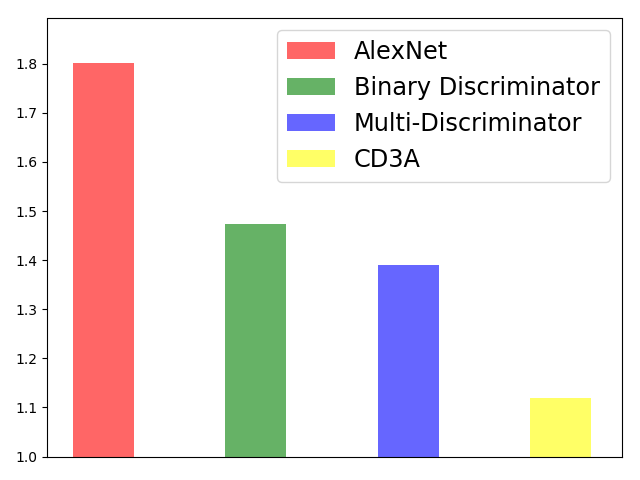}
     & \includegraphics[scale=0.30, width=0.20 \linewidth ,height=0.10 \textheight]{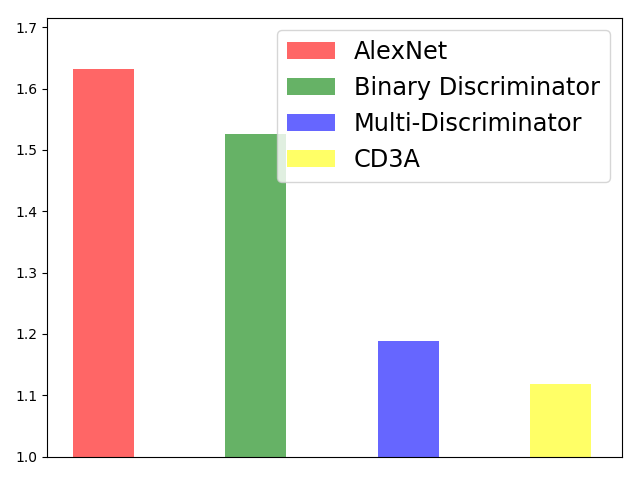}
       \end{tabular}
      \caption{(a) and (b) figures show t-SNE visualizations of the CNN's activation (a) in case when adapted through\cite{ganin_ICML2015} and (b) when adapted through proposed model. Blue and red points correspond to the source domain(A), and the target domain (W) respectively. Sub figures (c) and (d) show the proxy distance for A$\rightarrow$D and A$\rightarrow$W.}
      \label{tbl:tSNE}

 \end{figure*}
\subsection{Analysis}
\begin{figure*}[!]
 \small
\centering
 \begin{tabular}[b]{ c   c }
(a) CD$^3$A v/s D$^3$A Model  A$\rightarrow$W &  (b) SSA Plot for A$\rightarrow$W   \\
\includegraphics[height=0.18\textheight,width=0.65\textwidth]{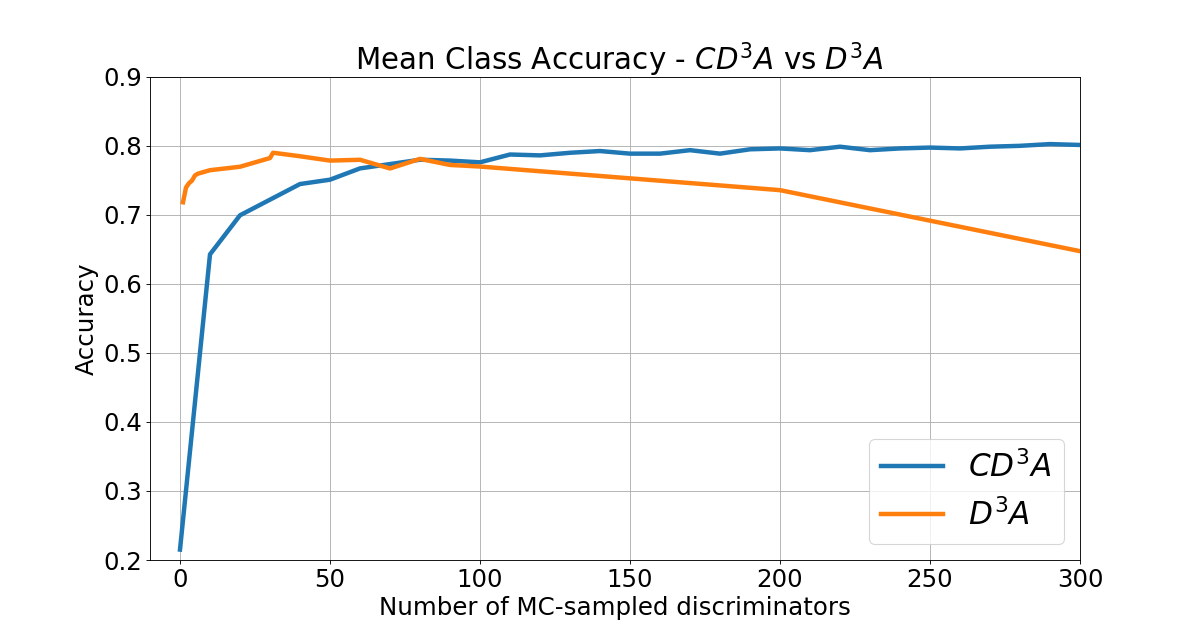} & 
\begin{minipage}[b][0.18\textheight][s]{0.30\textwidth}
  \centering
  \includegraphics[height=0.08\textheight,width=\textwidth]{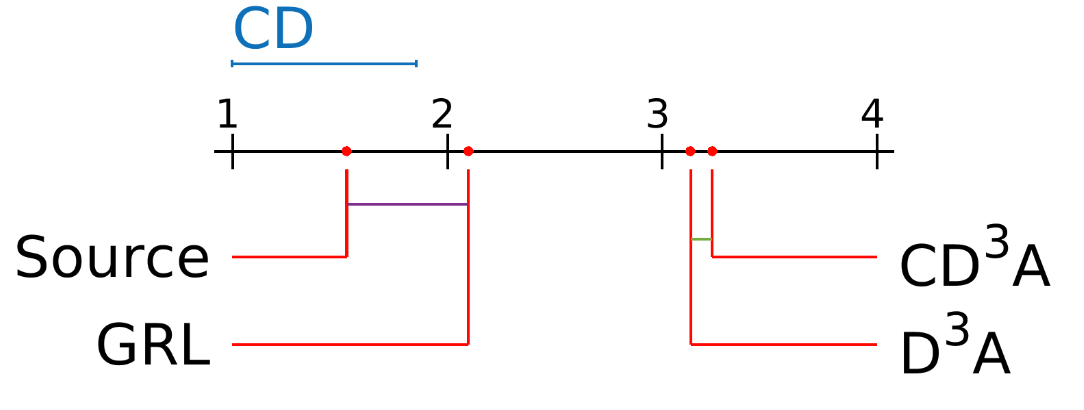}
\end{minipage}
  \end{tabular}
  
\caption{(a) Accuracy v/s Number of MC samples for D$^3$A and CD$^3$A model. Note that in  D$^3$A model, each model is trained separately and reported accuracy after the training, while in  CD$^3$A model the accuracy is calculated with single training process (b) Analysis of statistically significant difference for A $\rightarrow$ W in Binary label Discriminator (GRL) \cite{ganin_ICML2015}, proposed model and Source only methods, with a significance level of 0.05.}
\label{fig:ssa}
\end{figure*}

\textbf{{Curriculum v/s Fixed sampling:}}
We have plotted the accuracy as a function of the number of MC samples for both the models, curriculum-based sampling (CD$^3$A) and fixed sampling\newline (D$^{3}$A) in Figure~\ref{fig:ssa} (a). We can clearly observe that in the case of D$^{3}$A, the performance increases as we increase the number of MC sampled discriminators, but after some samples, the performance starts to deteriorate. While in case of CD$^3$A, the performance of model saturates after certain epochs. We can also see that CD$^3$A outperforms D$^{3}$A.
\newline
\textbf{Model complexity comparison with MADA:}
The proposed CD$^3$A model uses one discriminator(ensemble using dropout) whereas MADA uses as many discriminators as are the number of classes. Therefore, CD$^3$A has very few parameters as compared to MADA even for datasets with a small number of classes. For instance, in case of Office-31 dataset, MADA has 31 discriminators compared to CD$^3$A, which has only one discriminator. MADA has $\sim$98M parameters, while CD$^3$A has $\sim$59M parameters for Office-31 dataset. If we further increase the class size, the number of parameters in MADA increases(by $\sim$1.3M for every class label), but  CD$^3$A will have constant number of parameters ($\sim$59M). 
\newline
\textbf{{Feature visualization:}}
The adaptability  of target to source features can be visualized using the t-SNE embeddings of image features. We follow similar setting as in~\cite{ganin_ICML2015} to plot t-SNE embeddings for A$\rightarrow$W adaptation task in Figure~\ref{tbl:tSNE} (a) and (b). From the plot, we observe that adapted features(CD$^{3}$A) are more domain invariant than the features adapted with GRL.


\noindent\textbf{{Statistical significance analysis:}}
We analyzed statistical significance~\cite{demvsar_JMLR2006} for our C$D^{3}$A model against GRL\cite{ganin_ICML2015} and source only method for the domain adaptation task. The Critical Difference (CD) for Nemenyi test depends upon the given confidence level (0.05 in our case) for average ranks and  number of tested datasets. If the difference in the rank of the two methods lies within CD (our case CD = 0.6051), then they are not significantly different. Figure~\ref{fig:ssa}(b) visualizes the posthoc analysis using the CD diagram for A$\rightarrow$W. From the figures, it is clear that our C$D^{3}$A model is better and significantly different from other methods.



\noindent\textbf{{Proxy-$\mathcal{A}$- Distance}}
  $\mathcal{A}$-distance as a measure of cross domain discrepancy\cite{ben_ML2010}, which, together with the source risk, will bound the target risk. The proxy $\mathcal{A}$-distance is defined as $d_\mathcal{A} = 2 (1 - 2\epsilon)$, where $ \epsilon $ is the generalization error of a classifier(e.g. kernel SVM) trained on the binary task of discriminating source and target. Figure~\ref{tbl:tSNE}(c) and (d) shows $d_\mathcal{A}$ on tasks A $\rightarrow $D and  A $\rightarrow $W, with features of source only model\cite{krizhevsky_NIPS2012}, GRL\cite{ganin_ICML2015}, MADA\cite{pei_arxiv2018} and proposed model C$D^{3}$A. We observe that $d_\mathcal{A}$ calculated using C$D^{3}$A model features is much smaller than calculated using source only model, GRL and MADA features, which suggests that representations learned via C$D^{3}$A can reduce the cross-domain gap more effectively. 
\section{Conclusion}
In this paper, we provide a simple approach to obtain an improved discriminator for adversarial domain adaptation. We specifically show that the use of sampling-based ensemble results in an improved discriminator without increasing the number of parameters. The main reason for this improvement is that the features are made domain invariant based on a distribution of observations as against a single point estimate. Our approach based on curriculum dropout suggests that we are able to obtain an improved discriminator that is stable and improves the feature invariance learnt. We compare our method with standard baselines and provide a thorough empirical analysis of the method. We further observe through visualization that domain adapted features do result in domain invariant feature representations. 
Using the discriminator obtained through curriculum based dropout to solve domain adaptation is a promising direction, which we have initiated through this work.

\textbf{Acknowledgment}:  We acknowledge the resource support from Delta Lab, IIT Kanpur. Vinod Kurmi acknowledges support from TCS Research Scholarship Program.
\newpage

\bibliography{egbib}
\end{document}